\documentclass[10pt,twocolumn,letterpaper]{article}

\usepackage{cvpr}              
\usepackage{amsmath,amssymb,amsfonts}
\usepackage{mathtools}
\usepackage{graphicx}
\usepackage{textcomp}
\usepackage{algorithm}
\usepackage[noend]{algpseudocode}
\usepackage{booktabs}
\usepackage{multirow}
\newcommand*\numcircledmod[1]{\raisebox{.5pt}{\textcircled{\raisebox{-.9pt} {#1}}}}
\definecolor{cvprblue}{rgb}{0.21,0.49,0.74}
\usepackage[pagebackref,breaklinks,colorlinks,allcolors=cvprblue]{hyperref}

\def\paperID{*****} 
\def\confName{CVPR}
\def\confYear{2026}

\title{Exploring Budgeted Image Classification with Content-Sensitive Resource Allocation}

\author{Athanasios G. Papadopoulos\\
Tandon School of Engineering, New York University\\
{\tt\small tpapadop@nyu.edu}
}

\begin{document}
\maketitle
\begin{abstract}
The ever-growing adoption of Artificial Intelligence (AI) creates the need to deploy Deep Neural Networks in a variety of computational environments.
We consider dynamic environments, where computational requirements are subject to change, and we pose the following question:
How do we adjust the complexity of an AI classification system,
in order to maximize its accuracy, while meeting changing computational constraints?
We call this problem Budgeted Image Classification, and we formally formulate it as a resource allocation integer program. Given a computational budget, a batch of images, and a classification system that can make decisions with varying complexity (it has multiple decision points), we explore strategies to allocate images to decision points, in order to maximize accuracy within the available budget.
The original integer program is NP-Hard, so, we propose a continuous relaxation, leading to a content-agnostic allocation strategy
which assigns images to decision points without considering their particular content.
We address this issue by proposing a content-sensitive strategy, that we experimentally show it leads to superior performance.
We theoretically study the behavior of our strategies, deriving conditions that must be satisfied by decision points to be suitable for budgeted classification. We analyze fails cases, offering insights for future research directions.
\end{abstract}    
\section{Introduction}
\label{sec:introduction}
The widespread use of Artificial Intelligence (AI)
creates the need for training and deployment of deep models in a wide range of computational environments,
which vary from highly distributed data centers, to mobile phone devices~\cite{goyal2017accurate, howard2017mobilenets, howard2019searching}.
A straightforward way to address this problem, is to design families of models that have different computational complexity, and are suitable for different environments~\cite{tan2019efficientnet, tan2021efficientnetv2, bello2021revisiting, touvron2021training}.
This has become a trend in generative AI, where models like GPT and LLaMA are released in a range of sizes~\cite{brown2020language, touvron2023llama}.
However, computational environments may be dynamic, meaning that requirements may not be fixed, but subject to change.
This can be the case when a mobile phone enters to energy-saving mode, or its processor experiences high utilization and available resources are scarce.
As another example, computational requirements may change due to financial or environmental policies, since processing load is connected to monetary and environmental costs~\cite{strubell2019energy}.

In such scenarios, we would like an AI system to adjust its processing in such a way, that the new computational requirements are respected, while performance is maximized.
This assumes that our system has the ability to adjust its processing, and that we have a resource allocation strategy which dictates how to adjust the processing for optimal performance.
For example, given an ensemble of image classification models with different sizes, a computational budget $B$, and a batch of images $I$, we would like to determine how to allocate the images across the models of the ensemble, in order to not exceed $B$, and maximize accuracy in $I$. It is fair to assume that the smaller the size of a model, the less accurate it is expected to be, so, in an ideal scenario, we would classify all images with the largest model available. However, our budget $B$ may not allow for this, so we need to decide which images will be assigned to which model.

In this work, we focus on the problem of budgeted image classification, since classification is one of the most general and common computer vision tasks.
We generalize the concept of adjustable processing beyond model ensembles, by considering any processing system with multiple decision points.
Decision points (DPs) correspond to different processing paths that lead to a classification decision, and vary in computational cost and expected accuracy.
In the previous example, each model of the ensemble corresponds to a different decision point. We experiment with two types of adjustable processing systems, an ensemble of models that operate on inputs of different resolution, and a hard-attention model which attends to a varying number of image regions.
Our goal is to use these adjustable classification systems to explore the behavior of different allocation strategies, meaning different ways to distribute images to decision points under various computational budgets. To this end, we develop a content-agnostic, and a content-sensitive strategy.

The content-agnostic strategy determines how many images should be assigned to each DP in order to respect the available budget, but it doesn't specify which images should be assigned to each DP.
We experimentally demonstrate that it leads to suboptimal performance.
The development of our content-sensitive strategy is based on the simple, but important observation, that correct classification of certain images may be harder than others, e.g., due to partial occlusion of a salient object.
Based on that, easier images can be assigned to less accurate and less expensive DPs, while more difficult images can be assigned to more expensive, but more accurate DPs. Following this principle, our content-sensitive strategy estimates the difficulty of correctly classifying different images, and assigns them to DPs according to their accuracy-computational efficiency trade-off.

We formally formulate budgeted image classification as a resource allocation optimization problem, we develop two different allocation strategies to approximate its optimal solution, and we study their behavior, both experimentally and theoretically. Our goal is not to develop a state-of-the-art solution for a particular setting of budgeted classification, instead, we aim to provide a general formal view of the problem, and offer insights about the behavior of potential solutions.
The contributions of our work are the following:
\begin{itemize}
  \item We formulate budgeted image classification as an integer program, which is NP-Hard. We offer a continuous relaxation of the program to derive a tractable, but content-agnostic, solution. We enhance the content-agnostic solution with content-sensitive heuristics to provide a better approximation to the solution of the original integer program.
  \item We use two different adjustable processing systems, a multi-resolution ensemble and a hard-attention model, to experimentally demonstrate the performance benefits of our content-sensitive approximate solution.
  \item We study the conditions that must be satisfied by DPs in terms of their accuracy-computational efficiency trade-off, in order to be suitable for budgeted classification.
  \item We experimentally demonstrate limitations of content-sensitive heuristics. Importantly, we show that more expensive and more accurate DPs may not always be the optimal allocation choice, due to their specialization on different modes of the data distribution.
\end{itemize}
\section{Related Work}
\label{sec:related_work}
Budgeted classification requires an AI system with multiple decision points of different complexity, to accommodate adjustable processing.
Existing solutions include ensembles of networks with different inference costs~\cite{bolukbasi2017adaptive, ruiz2019adaptative},
models with classification heads attached to layers of different depth~\cite{huang2017multi, bolukbasi2017adaptive, teerapittayanon2016branchynet, li2019improved},
models that dynamically change their depth~\cite{figurnov2017spatially, veit2018convolutional, wang2018skipnet} or width~\cite{yu2018slimmable, gao2018dynamic},
ensembles of models that process inputs of different size~\cite{yang2020resolution, wang2021not},
models that employ various gating mechanisms to create processing paths of different cost~\cite{odena2017changing, shazeer2017outrageously, graves2016adaptive, verelst2020dynamic},
and hard-attention models which take sequences of glimpses of variable length~\cite{papadopoulos2021hard, li2017dynamic, wang2020glance}.

In addition to adjustable processing, budgeted classification requires an allocation mechanism that decides how to distribute images among the decision points.
A popular approach is to create an allocation strategy with trainable parameters, that learns to make decisions based on the content of individual images~\cite{figurnov2017spatially, verelst2020dynamic, bolukbasi2017adaptive, li2017dynamic}.
For example, in networks with decision points at layers of different depth, the allocation mechanism consists of binary functions that decide after each convolutional block whether processing should stop, or continue~\cite{bolukbasi2017adaptive}.
In such cases, constraints on the budget are formulated as loss terms, leaving budget changes to require retraining of the allocation strategy.
Trainable allocation methods can avoid retraining, by receiving computational requirements as an additional input during inference~\cite{odena2017changing}.
However, computational requirements are provided in the form of proxy variables, which don't have an obvious correspondence with the actual budget measured, e.g., in floating point operations (FLOPs).

Allocation strategies can also be non-trainable~\cite{huang2017multi, wang2021not, wang2020glance, yang2020resolution}.
For example, the given budget can be used to set up an optimization problem, which calculates a threshold value for every decision point.
During inference, processing stops at the first decision point that makes a prediction with confidence score (e.g., softmax probability) above the corresponding threshold~\cite{wang2021not}.
A problem with such methods is that threshold values are specified based on a validation set, thus, do not guarantee that the budget constraint will be satisfied for the batch at hand.
In addition, when the budget changes, a new optimization problem has to be solved, and its complexity may be a factor to consider.
The allocation strategies that we follow are non-trainable, and we solve an optimization problem as well, though, we guarantee that the budget is respected, and the optimization complexity is polynomial to the practically small number of decision points.
\section{Models}
\label{sec:background_information}
We use two different types of image classification approaches, which both have the ability to make predictions at multiple decision points, with varying computational requirements.
The first approach is an ensemble of models which share the same architecture, but process images at different resolution.
The second approach is a hard attention-model, TNet~\cite{papadopoulos2021hard},
which gradually accumulates information from an image, by selectively processing patches in a coarse-to-fine fashion.

\subsection{Multi-Resolution Ensemble}
\label{sec:efficientnets}
The ensemble consists of $3$ EfficientNet-B0 models~\cite{tan2019efficientnet}, which are trained independently on images of different resolutions. We denote the $3$ models as EN-B$0$-$224$, EN-B$0$-$448$, and EN-B$0$-$896$, because they are trained on images of resolution $224 \times 224$ px, $448 \times 448$ px, and $896 \times 896$ px, respectively.
Each model is a separate decision point, with different computational complexity due to the different input size.
We can think of the ensemble as processing an image pyramid, where EN-B$0$-$224$ is the least expensive DP with the lowest expected accuracy, while EN-B$0$-$896$ is the most expensive DP, with the highest expected accuracy.

\subsection{Traversal Network (TNet)}
\label{sec:tnet}
\begin{figure}[!t]
\centering
\includegraphics[width=0.45\textwidth]{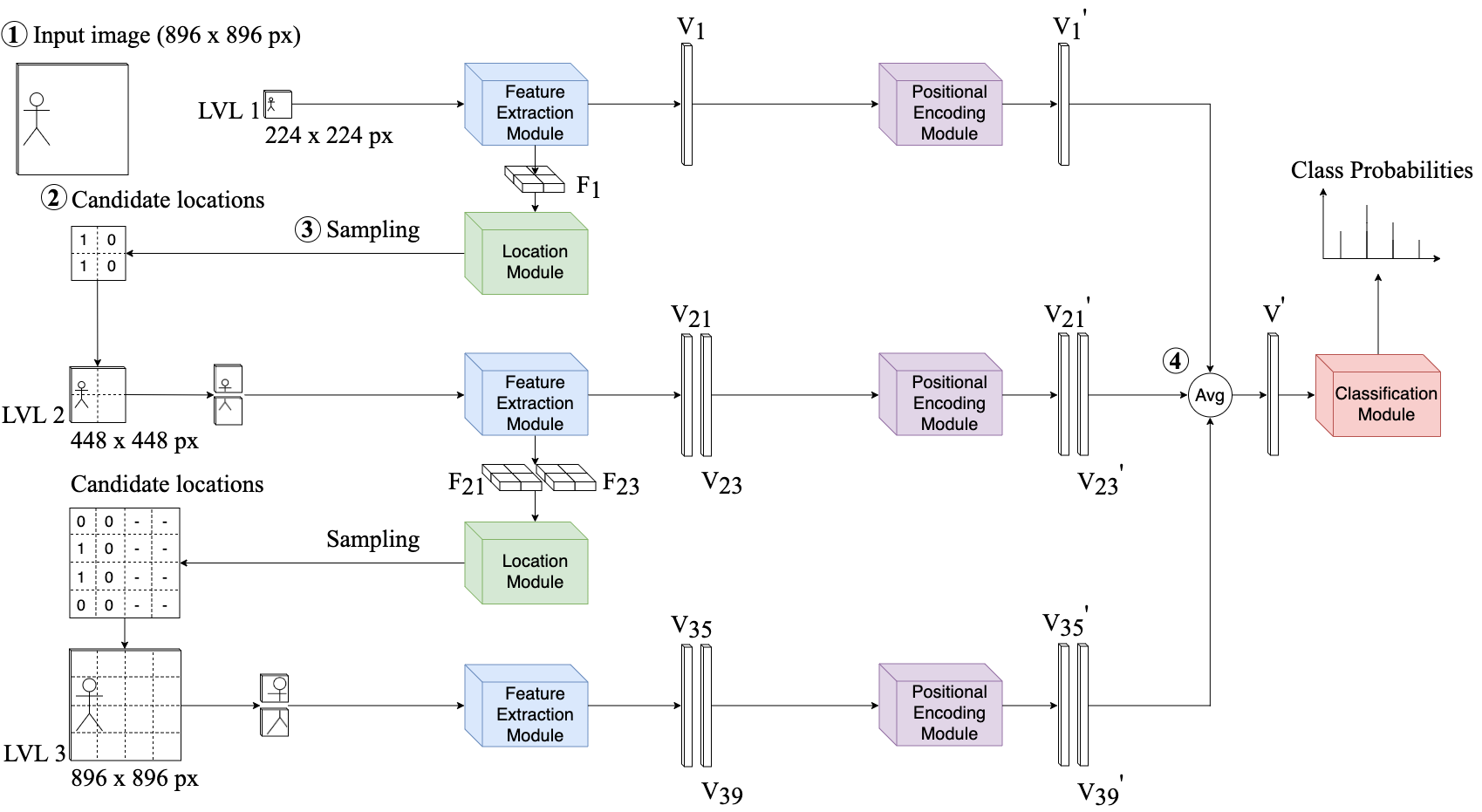}
\caption{Three unrolled processing levels of TNet~\cite{papadopoulos2021hard}. Starting at level~$1$, the image is processed in the coarsest scale (\emph{Feature Extraction Module}), and the extracted features are used to decide which image regions should be processed in a finer scale (\emph{Location Module}). This process is repeated for each selected location to reach level $3$, where features from the finest scale are extracted. All features are enriched with positional information
(\emph{Positional Encoding Module}), and then are averaged before the final classification (\emph{Classification Module}). Modules of the same name (and color) share weights.}
\label{tnet_arch}
\end{figure}

TNet is a multiscale hard-attention architecture, inspired by the way humans use saccades to explore the visual world~\cite{papadopoulos2021hard}.
Processing evolves in levels, where at each level, higher-resolution details become available to the model. In this work, we use TNet to process images of resolution $896 \times 896$ px.
We provide an overview of the TNet architecture in Figure~\ref{tnet_arch}.
Processing starts at level $1$ (\numcircledmod{1}), where we down-scale the full-resolution input to $224 \times 224$ px, and pass it through the \emph{feature extraction module}, to extract a feature vector $V_1$ that contains a coarse description of the original image.

To proceed to the next level, we feed an intermediate feature map, $F_{1}$, from the feature extraction module, to the \emph{location module}.
The location module considers a number of image regions as candidates to be processed in higher resolution, and predicts their importance.
In the example of Fig.~\ref{tnet_arch}, the candidate locations form a $2 \times 2$ regular grid (\numcircledmod{2}), and we sample $2$ locations (\numcircledmod{3}) based on their estimated importance.

In the 2nd processing level, we crop the selected regions from the full-resolution image, resize them to $224 \times 224$ px, and feed them to the feature extraction module to obtain the corresponding feature vectors, $V_{21}$ and $V_{23}$.
Then, we feed $F_{21}$ and $F_{23}$ to the location module, leading to $2$ new sampling grids. We sample $1$ location from each one of them to get $V_{35}$ and $V_{39}$ in the $3$rd processing level. 

Features extracted from all levels are passed through a \emph{positional encoding module}, which injects information about the spatial position and scale of the image regions.
We average all extracted features, $\lbrace V^{'}_{*} \rbrace$, into a single comprehensive representation, $V^{'}$ (\numcircledmod{4}), and we feed it to the \emph{classification module} for the final prediction.

TNet is flexible, and can be used to make predictions after each processing level, and with a varying number of attended locations.
This way, we can define multiple decision points, where DPs with more attended locations are expected to be more accurate, but with higher computational cost.

In this work, we use an EfficientNet-B0 backbone as the feature extraction module of TNet. The rest of the modules are defined according to~\cite{papadopoulos2021hard}; more details are provided in Section~\ref{sec:data}.
\section{Methods}
\label{sec:methods}
We define the problem of budgeted image classification in the following way:
Given a classification system $M$ with $K$ decision points of varying computational complexity, a computational budget $B$, and a batch of $N$ images $I = \{x_i | i=1, 2,...,N\}$, we want to maximize the classification accuracy in the batch, while the total computational cost does not exceed $B$.
Different metrics can be used for $B$. Without loss of generality, we use floating point operations (FLOPs), and holds $B \in \mathbb{R}^{+}$.

\subsection{Budgeted Image Classification Formulation}
\label{sec:resource_allocation}
We approach budgeted image classification as a resource allocation problem, and we seek its solution by solving the following integer program:
\begin{subequations} \label{eq16}
\begin{align}
\tag{\ref{eq16}}
\min_p \; &\frac{1}{N} \sum_{i=1}^{N} \sum_{k=1}^{K}{p_k^i \cdot E_k^i}, \\
\label{eq17}
\textnormal{subject to } &\sum_{i=1}^{N} \sum_{k=1}^{K}{p_k^i \cdot C_k} \leq B, \\
\label{eq18}
&\sum_{k=1}^{K}{p_k^i} = 1, \; \forall i \in \{1,2,...,N\}, \\
\label{eq19}
&p_k^i \in \{0, 1\}, \forall i \in \{1,2,...,N\} \wedge \forall k \in \{1,2,...,K\}
\end{align}
\end{subequations}

\noindent
where
$p_k^i$ is a binary variable that is equal to $1$ when image $x_i$ is classified at decision point $k$, and $0$ otherwise.
$p = \{p_k^i | i=1,2,...,N, k=1,2,...,K\}$ is the set of all $p_k^i$, which constitute our optimization variables.
$p$ represents allocation decisions about all $N$ images,
and constraints~\ref{eq18} and \ref{eq19}, guarantee that $p$ assigns each image to exactly one decision point.
$E_k^i$ is the probability of misclassifying image $x_i$ at decision point $k$.
The quantity minimized in expression~\ref{eq16}, is the expected classification error based on allocation $p$.
$C_k$ is the computational cost of classifying $x_i$ at decision point $k$; we assume that $C_k$ does not depend on image $x_i$.
Inequality~\ref{eq17}, requires the total classification cost that stems from allocation $p$, to not be greater than $B$.

The solution to program~\ref{eq16} is optimal allocation $p^*$, which assigns images to DPs in such a way, that the average expected error is minimal within the available budget.
In general, at decision points with lower cost, $C_k$, we expect easier images to have lower misclassification probability, $E_k^i$, compared to harder ones.
As we move to more expensive decision points, we expect $E_k^i$ to drop for all images, but for harder ones the drop to be more significant, since for easier ones, $E_k^i$ was already low.
Based on that, we expect optimal allocation, $p^*$, to assign easier images to decision points of lower cost, and harder images to more costly ones.
We note that more costly DPs may not always be preferable, meaning that for certain images, DPs of lower cost may have the lowest expected error, $E_k^i$. Program~\ref{eq16} accounts for that, by having a separate estimation $E_k^i$ for each image across DPs.

Linear program~\ref{eq16} is NP-Hard, with $N \cdot K$ variables.
This means that batch size, $N$, could easily get big enough that we face a computational bottleneck.
In addition, it is not straightforward how to estimate $E_k^i$ for all images and all decision points, especially since we are in a budgeted classification setting where computational requirements are strictly defined.

\subsection{Content-Agnostic (Random) Resource Allocation}
\label{sec:content_agnostic}
We simplify program \ref{eq16}, in $2$ ways.
First, we relax integrity constraint~\ref{eq19}, transforming NP-Hard integer program~\ref{eq16} into a linear program, solvable in polynomial time.
Second, we simplify the estimation of misclassification probabilities $E_k^i$, by using a method agnostic to image content, that is solely based on prior information.
In particular, we process the images of a held-out validation set, and at each decision point $k$, we calculate the total classification error $E_k$.
Then, for every image $x_i$ in $I$, we set $E_k^i = E_k$.
Based on that, all images are identical for the purposes of resource allocation, and our $N \cdot K$ variables $p_k^i$, reduce to $K$ relaxed variables $p_k$.
We get:
\begin{subequations} \label{eq20}
\begin{align}
\tag{\ref{eq20}}
\min_{p} \; &\sum_{k=1}^{K}{p_k \cdot E_k}, \\
\label{eq21}
\textnormal{subject to } &N \sum_{k=1}^{K}{p_k \cdot C_k} \leq B, \\
\label{eq22}
&\sum_{k=1}^{K}{p_k} = 1, \\
\label{eq23}
&0 \leq p_k \leq 1, \forall k \in \{1,2,...,K\}
\end{align}
\end{subequations}

\noindent
where $p_k$ is the probability of classifying an image at decision point $k$, and $p = \{p_k | k=1,2,...,K\}$ is the set of all $p_k$.
As in program~\ref{eq16}, the objective function is the expected classification error, and constraint~\ref{eq21} enforces the total computational cost to not be greater than $B$.
Constraints~\ref{eq22} and \ref{eq23}, require $p$ to parametrize a categorical probability distribution over decision points.
The way we interpret $p_k$, is that approximately $n_k =p_k \cdot N$ images should be classified at decision point $k$.
Based on that, we can have a content-agnostic allocation method, which solves linear program~\ref{eq20}, and randomly assigns $n_k =\lfloor p_k \cdot N \rfloor$ images\footnote{If $n_r = N - \sum_{k=1}^{K}{n_k} > 0$, we assign the remaining $n_r$ images to the least expensive decision point, for which holds $n_k > 0$. This way, we ensure that constraint~\ref{eq21} is satisfied.} to each decision point $k$.
We refer to this strategy either as \emph{content-agnostic allocation}, or \emph{random allocation}.

\subsection{Content-Sensitive Resource Allocation}
\label{sec:content_sensitive}

\begin{figure}[!t]
\begin{center}
\centerline{\includegraphics[width=0.5\columnwidth]{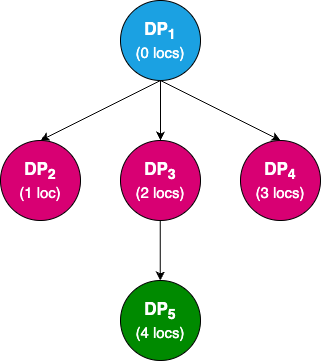}}
\caption{Processing tree of the TNet model (see Section~\ref{sec:tnet}).
The root node, $DP_1$, corresponds to the decision point at the $1$st processing level, which uses the down-scaled version of the whole image ($0$ attended locations).
The nodes at the next tree level can be executed in parallel, and correspond to decision points from TNet's $2$nd processing level; $DP_2$ corresponds to attending to $1$ location, $DP_3$ to $2$ locations, and $DP_4$ to $3$ locations.
$DP_5$ corresponds to a classification decision at TNet's $3$rd processing level, which is made by attending to $4$ locations; $2$ locations in the $2$nd processing level, and $1$ additional location within each one of them, in the $3$rd processing level.
}
\label{fig_dtree_tnet}
\end{center}
\end{figure}

We would like the decisions of our allocation strategy to be sensitive to image content.
To this end, we solve linear program~\ref{eq20} to get content-agnostic allocation probabilities $p = \{p_k | k = 1, 2, ..., K\}$, and we calculate the number $n_k$ of images that should be classified at each decision point $k$, as we described in Section~\ref{sec:content_agnostic}.
Then, we make content-sensitive decisions about which images should be classified at each decision point.

Decision points can be executed either in parallel or sequentially.
We represent processing dependencies between decision points by using a tree structure that we call \emph{processing tree}.
Each node in a processing tree corresponds to a different decision point, and for simplicity, we use the abbreviation DP to refer to both decision points and tree nodes.
Nodes at a tree level, $l$, can be executed in parallel, but they have to be executed after parent nodes at level $l - 1$, and before children nodes at level $l + 1$.

As an example, in Figure~\ref{fig_dtree_tnet}, we depict a processing tree of the TNet model (see Section \ref{sec:tnet}).
The root node ($DP_1$) corresponds to the classification decision that is made at the $1$st processing level,
based on the down-scaled version of the whole image.
The nodes at the next tree level correspond to decision points from TNet's $2$nd processing level.
In particular, $DP_2$ corresponds to attending to $1$ location, $DP_3$ to $2$ locations, and $DP_4$ to $3$ locations.
There could be more decision points, since more locations could be attended, but the current processing tree is sufficient for our purposes.
$DP_5$ represents a classification decision at TNet's $3$rd processing level, which is made by attending to $4$ locations.
Its parent node is $DP_3$ because the $4$ attended locations result from attending to $2$ locations in the $2$nd processing level, and to $1$ additional location within each one of them, in the $3$rd level.

A processing tree can be derived from any adjustable processing method, and it is at the core of our content-sensitive allocation strategy.
In particular, our strategy is based on the following simple idea. When decision points are executed sequentially, a parent DP should process all images that may be assigned to its children DPs, and forward to them the images it is more uncertain about. That's a heuristic to assign harder images (higher uncertainty) to more accurate DPs (children DPs are more expensive than their parents, so they are part of the solution to program~\ref{eq20} only if they are more accurate on average; lower $E_k$).

Our strategy is summarized in Algorithm~\ref{alg_alloc}. We use the processing tree of Fig.~\ref{fig_dtree_tnet} as an example to walk through the allocation algorithm.
We remind that we first solve linear program \ref{eq20} to calculate the number $n_k$ of images that should be classified at each $DP_k$ (lines 3-6 in Algorithm~\ref{alg_alloc}).
Then, we make a random initial assignment of images to DPs. In particular, for each $DP_k$ we create a set $I_k$ that contains $n_k$ randomly selected images from $I$; each image in $I$ is assigned to exactly one set $I_k$ (lines 7-8 in Algorithm~\ref{alg_alloc}).

Given processing tree $T$, we iterate over DPs in a breadth-first fashion, in order to update assignments $I_k$, and make final predictions $P$ (lines 9-18 in Algorithm~\ref{alg_alloc}).
At each $DP_k$, we process all images assigned to DPs in subtree $T_k$;
$T_k$ is the subtree of $T$ with $DP_k$ as its root.
We also define $I_{T_k}$ as the set of all images assigned to DPs that are part of $T_k$.
For example, given the processing tree of Fig.~\ref{fig_dtree_tnet}, if we are at $DP_1$, subtree $T_1$ is equivalent to the whole tree $T$, and we process all $N$ images in $I$; $T_1 \equiv T$, and $I_{T_1} = I$. If we are at $DP_3$, $T_3$ consists of $DP_3$ and $DP_5$, $I_{T_3} = I_3 \cup I_5$, and we process all $n_3 + n_5$ images in $I_{T_3}$ (lines 10-12 in Algorithm~\ref{alg_alloc}).

After all images in $I_{T_k}$ are processed at $DP_k$, they are sorted in decreasing order of top-1 classification probability; a content-based heuristic to sort images in order of increasing difficulty (line 13 in Algorithm~\ref{alg_alloc}).
Then, we iterate over every $DP_i$ in $T_k$ in order of increasing computational cost.
We update $I_i$ to contain the first $n_i$ sorted images in $I_{T_k}$, we remove these images from $I_{T_k}$, and we move to the next DP (lines 14-16 in Algorithm~\ref{alg_alloc}).
This is an attempt to assign images of greater difficulty, to more accurate DPs.

After we update $I_i$ for every $DP_i$ in $T_k$, we store at the set of final predictions, $P$, the predictions made by the root $DP_k$ for the $n_k$ images in $I_k$, and we move to the next DP
(line 17 in Algorithm~\ref{alg_alloc}).

To put everything together for the processing tree of Fig~\ref{fig_dtree_tnet}, we start at $DP_1$, where we process all $N$ images. Then, we assign the $n_1$ images with the highest top-1 classification probability to $DP_1$, the next $n_2$ images to $DP_2$, the next $n_3$ images to $DP_3$, the next $n_4$ images to $DP_4$, and the next $n_5$ images to $DP_5$.
We store the predictions made for the images in $I_1$ at the set of final predictions $P$, and we move to the next decision point, $DP_2$.
Subtree $T_2$ contains only $DP_2$, so, we just classify at $DP_2$ the images in $I_2$, and we store the predictions in $P$. We move to $DP_3$, where we process all images assigned to $I_3$ and $I_5$. We store the $n_3$ most confident predictions in $P$, and we assign the rest $n_5$ images to $I_5$.
$DP_4$ and $DP_5$ are the next two DPs to visit, where we just classify the already assigned images, and we store the predictions to conclude the process.

\begin{algorithm*}[!t]
\caption{Content-sensitive resource allocation.}
\label{alg_alloc}
\textbf{Input:} batch $I$ with $N$ images, budget $B$, system $M$ with $K$ decision points (DPs) and processing tree $T$, validation set $I_{vs}$, computational cost $C_k$ for every $DP_k$ \\
\textbf{Output:} classification predictions $P$ for the images in $I$
\begin{algorithmic}[1]
\For {$ k=1,2,...,K$} \Comment{offline processing}
\State $E_k \gets$ classification error on $I_{vs}$ at $DP_k$
\EndFor
\State $\{p_k | k=1,2,...,K\} \gets \text{solve linear program \ref{eq20}}$
\State $\{n_k | k=1,2,...,K\} \gets \lfloor p_k \cdot N \rfloor$
\State $r \gets$ find index $k$ of $DP_k$ with lowest $C_k$ for which holds $p_k > 0$ \Comment{ensure $\sum_{k=1}^K n_k = N$}
\State $n_r \gets n_r + N - \sum_{k=1}^{K}{n_k}$
\For {$k=1,2,...,K$} \Comment{initialize sets $I_k$ with images assigned to every $DP_k$}
\State $I_k \gets$ sample $n_k$ images without replacement from uniform distribution over $I$
\EndFor
\For {$DP_k$ $\in T$ in breadth-first order}
\State $T_k \gets$ subtree of $T$ with root $DP_k$
\State $I_{T_k} \gets$ $\cup_{i:DP_i \in T_k} I_i$ \Comment{$I_{T_k}$ is the set of all images currently assigned to DPs in $T_k$}
\State $probs \gets$ use $DP_k$ to calculate class probabilities of all images in $I_{T_k}$
\State $I_{T_k} \gets$ sort $I_{T_k}$ in decreasing order of top-$1$ probability in $probs$
\For {$DP_i$ $\in T_k$ in order of increasing cost $C_i$}
\State $I_i \gets$ overwrite $I_i$ with the first $n_i$ images of $I_{T_k}$
\State $I_{T_k} \gets$ remove the first $n_i$ images of $I_{T_k}$
\EndFor
\State $P \gets$ add predictions for the images in $I_k$
\EndFor
\State $\text{\textbf{return}} \; P$
\end{algorithmic}
\end{algorithm*}
\section{Experimental Evaluation}
\label{sec:experimental_evaluation}

\subsection{Data and Models}
\label{sec:data}
\begin{figure}[!t]
\begin{center}
\centerline{\includegraphics[width=0.8\columnwidth]{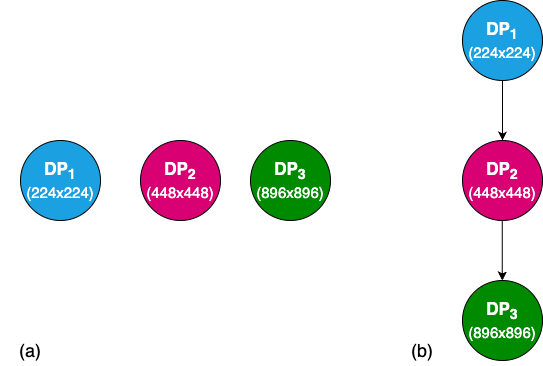}}
\caption{
Two processing trees of EN-B$0$ ensemble, which consists of EN-B$0$-$224$, EN-B$0$-$448$, and EN-B$0$-$896$ models (see Section~\ref{sec:efficientnets}).
The $3$ models constitute $3$ different decision points.
(a) In content-agnostic allocation, DPs are executed in parallel, since images are assigned randomly.
(b) In content-sensitive allocation, DPs are executed sequentially, in order of increasing computational cost; first EN-B$0$-$224$, then EN-B$0$-$448$, and last EN-B$0$-$896$.
}
\label{fig_dtree_ensemble}
\end{center}
\end{figure}

\begin{table}[!t]
\caption{
Expected misclassification probability and computational cost of every DP in the processing trees of EN-B$0$ ensemble.
For the content-agnostic case, the DPs can be executed in parallel, as depicted in Fig.~\ref{fig_dtree_ensemble}~(a), while for the content-sensitive case, the DPs are executed sequentially, as depicted in Fig.~\ref{fig_dtree_ensemble}~(b). The sequential execution leads $DP_2$ and $DP_3$ to have higher cost in content-sensitive allocation.
}
\label{cost_error_ensemble}
\begin{center}
\small
\setlength{\tabcolsep}{4pt}
\begin{tabular}{llccc}
\toprule
& & $\mathbf{DP_1}$ & $\mathbf{DP_2}$ & $\mathbf{DP_3}$ \\
\midrule
\multirow{2.5}{*}{\shortstack{\textbf{Content} \\ \textbf{Agnostic}}} & \textbf{Error}$\mathbf{-E_k}$ \textbf{(\%)} & $36.41$ & $29.39$ & $28.68$  \\
\cmidrule{2-5}
& \textbf{Cost}$\mathbf{-C_k}$ \textbf{(GFLOPs)} & $0.39$ & $1.54$ & $6.18$ \\
\midrule
\multirow{2.5}{*}{\shortstack{\textbf{Content} \\ \textbf{Sensitive}}} & \textbf{Error}$\mathbf{-E_k}$ \textbf{(\%)} & $36.41$ & $29.39$ & $28.68$  \\
\cmidrule{2-5}
& \textbf{Cost}$\mathbf{-C_k}$ \textbf{(GFLOPs)} & $0.39$ & $1.93$ & $8.11$ \\
\bottomrule
\end{tabular}
\end{center}
\end{table}

Functional Map of the World (fMoW)~\cite{fmow2018} consists of high-resolution satellite images, from $62$ classes.
They are split in $363,572$ training, $53,041$ validation and $53,473$ testing images.

For the multi-resolution ensemble of EN-B$0$-$224$, EN-B$0$-$448$, and EN-B$0$-$896$ (see Section~\ref{sec:efficientnets}), we use models trained on fMoW, provided in~\cite{papadopoulos2021hard}.
All members of the ensemble share the same architecture, and have $4.13$M parameters.

We consider two processing trees for EN-B$0$ ensemble, one for the case of content-agnostic (random) allocation, and one for content-sensitive allocation.
In both cases, the output of each model corresponds to a different decision point.
For content-agnostic allocation, images are assigned randomly to members of the ensemble, according to the solution to program~\ref{eq20}. As a result, decision points can be executed in parallel, leading to the processing tree of Figure~\ref{fig_dtree_ensemble}~(a).
For content-sensitive allocation, decision points are executed sequentially, in order of increasing computational cost; first EN-B$0$-$224$, then EN-B$0$-$448$, and last EN-B$0$-$896$, leading to the processing tree of Figure~\ref{fig_dtree_ensemble}~(b).
For both processing trees, the computational cost and expected misclassification probability of each DP are provided in Table~\ref{cost_error_ensemble}.
\begin{table}[!t]
\caption{
Expected misclassification probability and computational cost of every DP in the processing tree of TNet (Fig.~\ref{fig_dtree_tnet}).
}
\label{cost_error_tnet}
\begin{center}
\small
\setlength{\tabcolsep}{4pt}
\begin{tabular}{lccccc}
\toprule
& $\mathbf{DP_1}$ & $\mathbf{DP_2}$ & $\mathbf{DP_3}$ & $\mathbf{DP_4}$ & $\mathbf{DP_5}$ \\
\midrule
\textbf{Error}$\mathbf{-E_k}$ \textbf{(\%)} & $49.94$ & $29.48$ & $27.84$ & $27.7$ & $27.07$ \\
\cmidrule{1-6}
\textbf{Cost}$\mathbf{-C_k}$ \textbf{(GFLOPs)} & $0.39$ & $0.77$ & $1.16$ & $1.55$ & $1.94$ \\
\bottomrule
\end{tabular}
\end{center}
\end{table}

Similar to EN-B$0$ ensemble, we use the TNet model trained on fMoW that is provided in~\cite{papadopoulos2021hard}. As we described in Section~\ref{sec:tnet}, TNet's feature extraction module follows the EfficientNet-B$0$ architecture, and has a total number of $4.56$M parameters; the exact architectures of all modules can be found in~\cite{papadopoulos2021hard}.
The processing tree that we use in our experiments is depicted in Figure \ref{fig_dtree_tnet}, while the cost and expected misclassification probability of the corresponding DPs are provided in Table~\ref{cost_error_tnet}.

For each DP, we estimate its expected misclassification probability by calculating its average classification error on the validation set of fMoW. The required FLOPs are calculated according to~\cite{papadopoulos2021hard}.

\subsection{Results}
\label{sec:results}
We evaluate our models on the test set of fMoW, by using $2$ resource allocation methods.
We use the content-agnostic (random) allocation strategy described in Section~\ref{sec:content_agnostic}, and the content-sensitive allocation strategy presented in Section~\ref{sec:content_sensitive} and Algorithm~\ref{alg_alloc}.
For each classification approach (TNet and ensemble), we use $50$ different computational budgets, which uniformly cover the range $[B_{min}, B_{max}]$.
$B_{min}$ and $B_{max}$ are the computational costs of classifying all given images at the least, and most expensive decision point, respectively.

\subsubsection{Allocation Probabilities}
\label{sec:aloc_probs}
\begin{figure*}[!t]
\begin{center}
\centerline{\includegraphics[width=1.4\columnwidth]{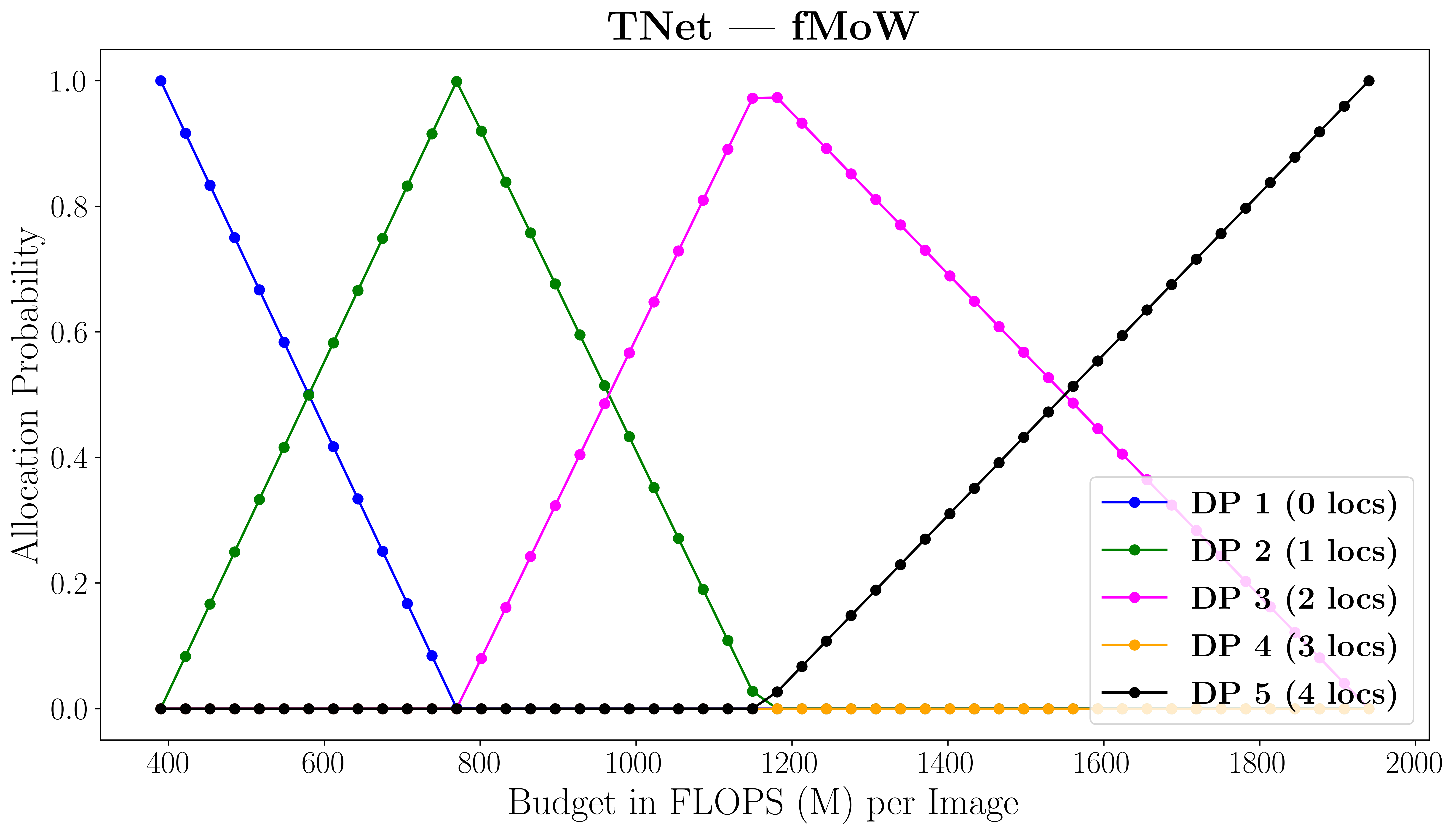}}
\caption{
Optimal allocation probabilities $p = \{p_k, \; k = 1, 2, ..., 5\}$ of TNet's decision points for the processing tree of Fig.~\ref{fig_dtree_tnet}, under different computational budgets.
We get $p$ by solving linear program~\ref{eq20}; expected error and cost of each DP is provided in Table~\ref{cost_error_tnet}.
For the lowest budget, all images are classified at DP~$1$ ($p_1 = 1$).
As the budget increases, $p_1$ starts to decrease, and $p_2$ to increase.
This corresponds to a general trend, where additional budget is used to classify images at more expensive decision points, since they are expected to be more accurate.
}
\label{fig_alloc_probs_tnet}
\end{center}
\end{figure*}

\begin{figure*}[!t]
\begin{center}
\centerline{\includegraphics[width=1.4\columnwidth]{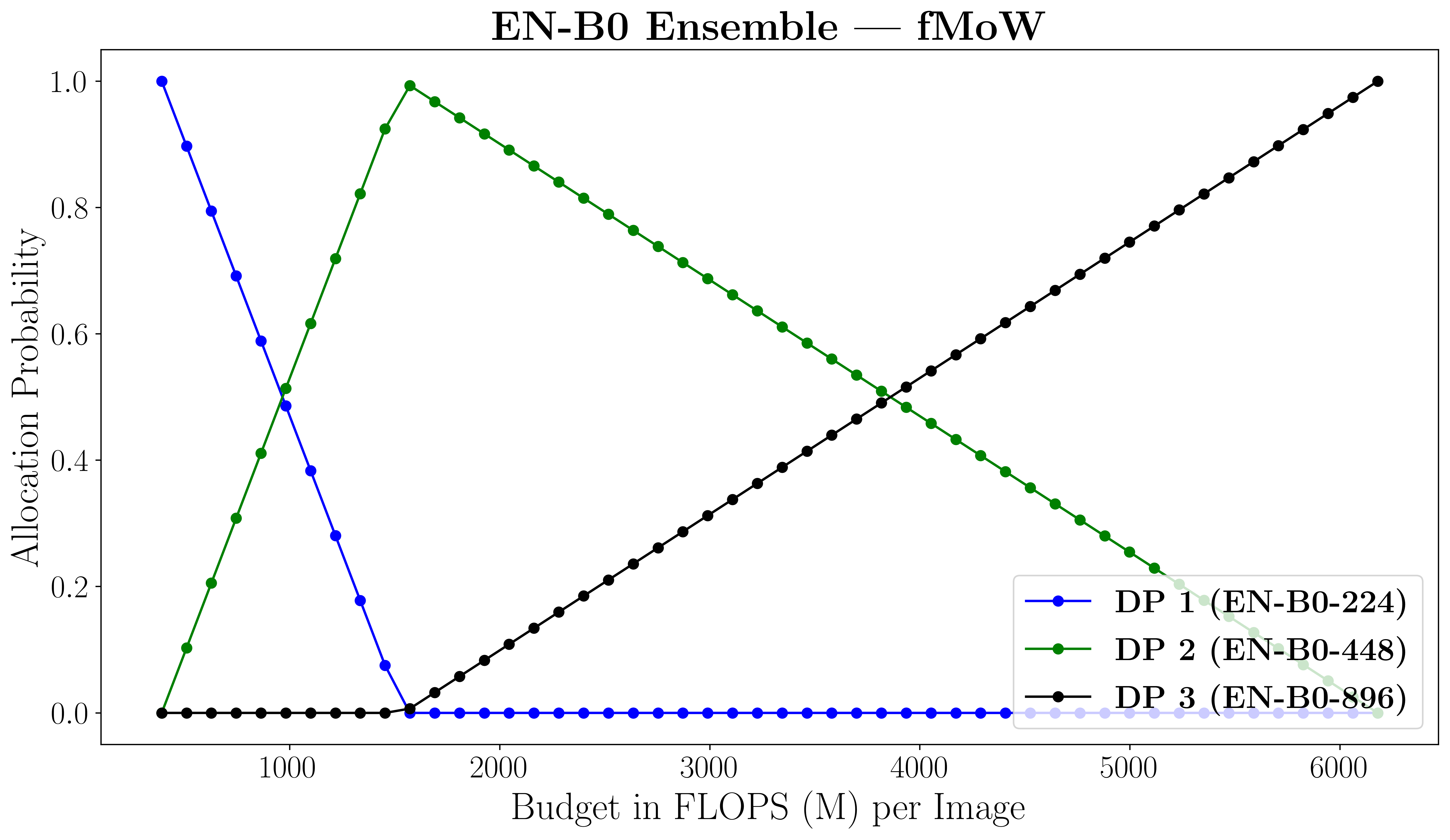}}
\caption{
Optimal allocation probabilities $p = \{p_k, \; k = 1, 2, 3\}$ of EN-B$0$ ensemble's decision points for the processing tree of Fig.~\ref{fig_dtree_ensemble}~(a), under different computational budgets.
We get $p$ by solving linear program~\ref{eq20}; expected error and cost of each DP is provided in Table~\ref{cost_error_ensemble}.
For the lowest budget, all images are classified at DP~$1$ ($p_1 = 1$).
As the budget increases, $p_1$ starts to decrease, and $p_2$ to increase.
This corresponds to a general trend, where additional budget is used to classify images at more expensive decision points, since they are expected to be more accurate.
For the content-sensitive case (Fig.~\ref{fig_dtree_ensemble}~(b)), the only difference is the cost of the DPs.
}
\label{fig_alloc_probs_bl}
\end{center}
\end{figure*}

Both content-agnostic and content-sensitive strategy require solving linear program~\ref{eq20}
to get allocation probabilities $p = \{p_k, \; k = 1, 2, ..., K\}$; for TNet holds $K=5$, and for the ensemble $K=3$.
In Figure~\ref{fig_alloc_probs_tnet}, we provide the allocation probabilities of TNet's $5$ decision points under different budgets.
For the minimum budget, $B_{min}$, all images are classified at $DP_1$ ($p_1=1$).
As the budget increases, $p_1$ starts to decrease, and $p_2$ to increase.
This means that the extra budget is used to classify more images at $DP_2$, instead of $DP_1$.
As the budget continues to increase, $p_1$ goes to $0$, and $p_2$ gets equal to $1$. Then, $p_2$ starts to decrease, and $p_3$ starts to increase.
The general behavior is that additional budget is used to classify images at more expensive decision points, since they are expected to be more accurate.
This is the behavior that we observe in Figure~\ref{fig_alloc_probs_bl} as well, where we provide the allocation probabilities of the EN-B$0$ ensemble under different budgets.

\subsubsection{Allocation Trade-Off}
\label{sec:trade_offs}
Interestingly, in Figure~\ref{fig_alloc_probs_tnet}, the probability of $DP_4$ remains equal to $0$ for all budgets. To understand why this is happening, we explore how budget changes affect the objective function of linear program~\ref{eq20}.

Let's assume that based on our budget $B$, the solution to program~\ref{eq20} is $p_r = 1$, meaning that we are classifying all images at $DP_r$. For example, this is the case when our budget is equal to $B_{min}$, and the allocation probability of the least expensive DP is equal to $1$.
To aid our analysis, we will call $DP_r$ the \textit{decision point of reference}.
Now, let's assume that an additional small budget $\Delta B > 0$ becomes available. Given $\Delta B$, we have the option to increase the allocation probability of any $DP_k$ that is more costly than $DP_r$ ($C_k > C_r$), while we have to decrease the allocation probability of the reference DP by the same amount
\footnote{For any $DP_k$ that holds $C_k \leq C_r$, if an increase in allocation probability $p_k$ could lead to a better solution to program~\ref{eq20}, $p_k$ would have been higher already, since the initial budget $B$ is sufficient to assign any number of images to DPs less costly than $DP_r$.}.
We get:
\begin{align}
\label{delta_pkr_1}
&\Delta p_{k,r} \cdot (C_k - C_r) = \Delta B \Rightarrow \\
\label{delta_pkr_2}
&\Delta p_{k,r} = \frac{\Delta B}{(C_k - C_r)}
\end{align}

\noindent
where $\Delta p_{k,r}$ is the maximum possible increase of $p_k$ when $DP_r$ is the decision point of reference. Equation~\ref{delta_pkr_1} indicates that the increase in total classification cost due to the increase of $p_k$ and decrease of $p_r$ (left-hand side), has to be equal to the budget increase $\Delta B$
\footnote{We assume that $\Delta B$ is small enough, such that $p_k \leq 1$ after it is increased by $\Delta p_{k,r}$, and $p_r \geq 0$ after it is decreased by $\Delta p_{k,r}$.
We further assume that $\Delta B$ can be used in its entirety.}.

The impact that the increase of $p_k$ will have on the objective function of program~\ref{eq20}, is the following:
\begin{align}
\label{delta_Ek_1}
&\Delta E_{k, r} = \Delta p_{k, r} \cdot (E_r - E_k) \Rightarrow \\
\label{delta_Ek_2}
&\Delta E_{k, r} = \Delta B \cdot \frac{E_r - E_k}{C_k - C_r}
\end{align}

\noindent
where $\Delta E_{k, r}$ is the change in total expected classification error when $p_k$ is increased by $\Delta p_{k, r}$ (and $p_r$ is decreased by the same amount). $E_k$ and $E_r$ are the expected misclassification probabilities at $DP_k$ and $DP_r$ respectively.
$\Delta E_{k, r}$ is positive when $DP_k$ is more accurate than $DP_r$ ($E_k < E_r$), and corresponds to the amount by which the objective function of program~\ref{eq20} will decrease.
The opposite holds for negative $\Delta E_{k, r}$, meaning that $DP_k$ has to be less accurate than $DP_r$, leading the objective function to increase by $| \Delta E_{k, r} |$.
Based on this, the optimal use of extra budget $\Delta B$, is to increase the allocation probability of $DP_k$ with the largest positive $\Delta E_{k, r}$:
\begin{subequations} \label{argmax_Ekr_1}
\begin{align}
\tag{\ref{argmax_Ekr_1}}
\underset{k}{\text{argmax}} \; &\Delta E_{k, r} = \underset{k}{\text{argmax}} \; \alpha_{k,r} \\
\label{argmax_Ekr_2}
\textnormal{s.t. } &\alpha_{k, r} > 0 \\
\label{argmax_Ekr_3}
\textnormal{where } &\alpha_{k,r} \coloneq \frac{E_r - E_k}{C_k - C_r}
\end{align}
\end{subequations}

\noindent
Equation~\ref{argmax_Ekr_1} follows from Eq.~\ref{delta_Ek_2}, by considering that the $\text{argmax}$ operation is independent of $\Delta B$. We introduce $\alpha_{k,r}$ to aid our analysis, and we call it the \textit{trade-off ratio} between $DP_k$ and $DP_r$, since it represents the ratio between the reduction in expected classification error ($E_r - E_k$), and the increase in computational cost ($C_k - C_r$), when the allocation probability $p_r$ is decreased in favor of $p_k$.

\begin{table}[!t]
\caption{Trade-off ratios, $\alpha_{k,r}$ (Eq.~\ref{argmax_Ekr_3}), of the TNet model, for $r \in [1, K-1]$, and $k \in [r+1, K]$, where $K = 5$.
DPs correspond to the processing tree of Fig. \ref{fig_dtree_tnet}.
Dashes, $-$, correspond to cases where either $C_k < C_r$, or $E_k > E_r$; as we note in Section \ref{sec:trade_offs}, these cases are ignored by the optimal solution to program~\ref{eq20} when additional budget becomes available. In bold is the largest value of each row. We observe that $DP_4$ never has the largest trade-off ratio, thus, it is ignored by our allocation strategies.
}
\label{trade_off_ratios}
\begin{center}
\begin{tabular}{cccccccc}
\toprule
\multirow{2.5}{*}{\textbf{Ref DP}} & \multicolumn{4}{c}{\textbf{Trade-off Ratio}} \\
\cmidrule{2-5}
& $\mathbf{DP_2}$ & $\mathbf{DP_3}$ & $\mathbf{DP_4}$ & $\mathbf{DP_5}$ \\
\midrule
$\mathbf{DP_1}$ & $\mathbf{53.84}$ & $28.7$ & $19.17$ & $14.75$  \\
$\mathbf{DP_2}$ & $-$ & $\mathbf{4.21}$ & $2.28$ & $2.06$ \\
$\mathbf{DP_3}$ & $-$ & $-$ & $0.36$ & $\mathbf{0.99}$ \\
$\mathbf{DP_4}$ & $-$ & $-$ & $-$ & $\mathbf{1.62}$ \\
\bottomrule
\end{tabular}
\end{center}
\end{table}

We can use problem~\ref{argmax_Ekr_1} to better understand how allocation probabilities change in Figure~\ref{fig_alloc_probs_tnet}. In particular, in Table~\ref{trade_off_ratios}, we calculate the trade-off ratio, $\alpha_{k,r}$, for $r \in [1, K-1]$, and $k \in [r+1, K]$. As we mentioned in Section~\ref{sec:aloc_probs}, the lowest available budget, $B_{min}$, corresponds to the case that every image is assigned to $DP_1$ ($p_1=1$). In the first row of Table~\ref{trade_off_ratios}, we use $DP_1$ as the decision point of reference, and we calculate the trade-off ratio for every other DP.
As we can see, $\alpha_{2,1}$ has the highest value, and as a result, the objective function of program~\ref{eq20} will be reduced the most when additional budget is used to increase $p_2$. This is what we see happening in Figure~\ref{fig_alloc_probs_tnet}, where $p_2$ gradually increases to $1$, as more budget becomes available.
When $p_2=1$, we can use $DP_2$ as the DP of reference, and based on the second row in Table~\ref{trade_off_ratios}, $\alpha_{3,2}$ has the highest positive value.
Similar to what we observed before, in Figure~\ref{fig_alloc_probs_tnet}, $p_3$ starts to increase until it gets equal to $1$, becoming the next DP of reference. In the third row of Table~\ref{trade_off_ratios}, we see that $\alpha_{5,3} > \alpha_{4,3} > 0$, thus, $DP_4$ is ignored, and $p_5$ starts to increase. Even if $DP_5$ is more costly than $DP_4$ ($C_5 > C_4$), meaning that we can assign to it fewer additional images ($\Delta p_{5,3} < \Delta p_{4,3}$ based on Eq.~\ref{delta_pkr_2}), the decrease in expected classification error is more significant ($E_5 < E_4$), making the expected reduction in the overall classification error greater (this is what is expressed by $\alpha_{5,3} > \alpha_{4,3}$). As a result, it is preferable to increase $p_5$ instead of $p_4$.

This explains why $p_4$ remains equal to $0$ under every budget in Fig.~\ref{fig_alloc_probs_tnet}. In general, we can use problem~\ref{argmax_Ekr_1} to find out in advance, if the DPs of our system are suitable for budgeted classification. In particular, we can calculate the trade-off ratios as in Table~\ref{trade_off_ratios}, and remove every DP that doesn't have the highest positive trade-off ratio in at least one row; these are the DPs that will never be part of the solution to program~\ref{eq20}.

\subsubsection{Classification Accuracy}
\label{sec:accuracy}
\begin{figure*}[!t]
\begin{center}
\centerline{\includegraphics[width=1.6\columnwidth]{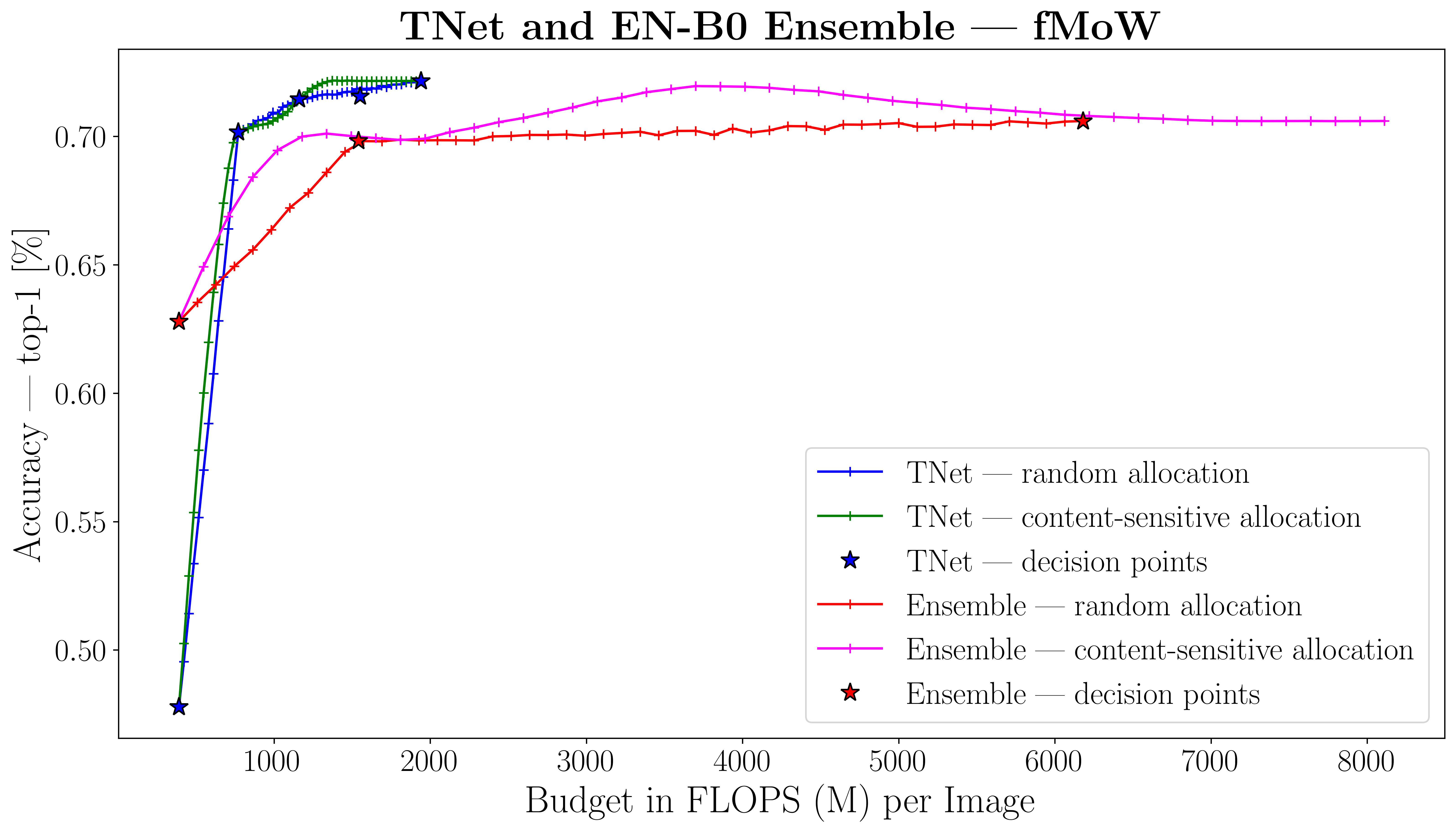}}
\caption{
Classification accuracy of TNet and EN-B$0$ ensemble on the test set of fMoW, under different computational budgets.
We use both the content-agnostic and the content-sensitive allocation methods.
The random allocation strategy leads to accuracy that increases linearly between decision points, while the content-sensitive strategy leads to higher accuracy under most budgets,
and in the case of EN-B$0$ ensemble, it even leads to considerably higher accuracy overall. 
}
\label{fig_alloc}
\end{center}
\end{figure*}

In Figure \ref{fig_alloc}, we provide the classification accuracy of TNet and EN-B$0$ ensemble on the test set of fMoW, under different computational budgets.
We use both the random and context-sensitive allocation strategy.
We are not interested in the performance differences between the ensemble and TNet, we rather focus on the behavior of each model under the two allocation strategies.

For both TNet and EN-B$0$ ensemble, the content-agnostic allocation strategy leads to accuracy that increases linearly between decision points (red and blue lines); except for TNet's $DP_4$, which constantly has $0$ allocation probability for reasons discussed in Section~\ref{sec:trade_offs}.
The content-sensitive allocation strategy leads both TNet and EN-B$0$ ensemble to higher accuracy under almost all budgets (magenta and green lines).

Its impact is more profound for EN-B$0$ ensemble, as we can see for example in the budget range between the ensemble's $DP_1$ and $DP_2$ (the two red stars of lower cost).
In this particular budget range, allocation probability $p_1$ decreases and $p_2$ increases, meaning that images that were previously classified at $DP_1$, are now classified at $DP_2$. Content-agnostic strategy randomly decides which images should be assigned to the two DPs, achieving classification error that is approximately equal to $p_1 \cdot E_1 + p_2 \cdot E_2$; a linear interpolation between $E_1$ and $E_2$.
In contrast, content-sensitive strategy assigns the easier images to $DP_1$ (the images that $DP_1$ is more confident about), and the harder images to the more accurate $DP_2$ (the images that $DP_1$ is less confident about).
This way, the classification error of $DP_1$ is lower than $E_1$ on the easier images, while the error of $DP_2$ remains close to $E_2$, leading to a significant improvement over the random allocation.

We note that the decision points of the ensemble are not executed sequentially during the content-agnostic allocation.
For example, when an image is classified at $DP_2$, the computational cost is equal to the cost of executing EN-B$0$-$448$ (processing tree of Fig.~\ref{fig_dtree_ensemble}~(a)).
In contrast, during content-sensitive allocation, when an image is classified at $DP_2$, the computational cost is equal to the cost of executing EN-B$0$-$224$ and EN-B$0$-$448$, since they are executed in a sequence (processing tree of Fig.~\ref{fig_dtree_ensemble}~(b)).
This is why the curve of the ensemble's content-sensitive allocation extends further along the budget axis, compared to the curve that corresponds to the content-agnostic allocation.
TNet follows the processing tree of Figure~\ref{fig_dtree_tnet} in both allocation strategies, and the cost of its decision points does not change.

\subsubsection{Specialization on Data Modes}
\label{sec:specialization}
Of particular interest is the behavior of the ensemble in the budget range between $DP_2$ and $DP_3$ (the two red stars of higher cost).
We see that the content-sensitive strategy reaches significantly higher accuracy compared to the random one, e.g., around budget of $4$ GFLOPs, but accuracy gradually drops as the allocation probability of $DP_3$ becomes equal to $1$.
We attribute this behavior to the fact that the sets of images that can be classified correctly by EN-B$0$-$448$ and EN-B$0$-$896$ overlap only partially.
In other words, we hypothesize that $DP_2$ and $DP_3$ specialize to different modes of the data distribution.

We test this hypothesis by calculating the joint accuracy between $DP_2$ and $DP_3$, meaning the accuracy when an image is correctly classified at either one of the two DPs. We find joint accuracy to be around $78 \%$ (error $22 \%$), which is considerably higher than the accuracy of either one of the two DPs (see Table~\ref{cost_error_ensemble}).
This shows that there are images that can be classified correctly by $DP_2$ and not by $DP_3$, while the opposite holds true as well.
The content-sensitive strategy predicts class probabilities for all images at $DP_2$, and assigns those with the highest top-$1$ probability to $DP_2$ because they are deemed easier, while the rest are assigned to $DP_3$.
It seems that this is an effective way to find images classified correctly by $DP_2$ and not by $DP_3$, because when the budget increases and images previously classified at $DP_2$ are assigned to $DP_3$, the total accuracy drops.

This specialization of members of EN-B$0$ ensemble is not entirely surprising, because they are trained on images of different resolution, thus, they have access to different kinds of information.
On the other hand, TNet gradually accumulates information by attending to more image regions, meaning that more costly DPs add to the features already extracted by the less costly ones, leading to a gradual improvement of image understanding.
\section{Limitations and Future Work}
\label{sec:limitations}
We identify a number of limitations in our work, and we consider potential future research directions to address them.
First, in Section~\ref{sec:content_agnostic} we set $E_k^i = E_k$, meaning that we made the simplifying assumption that every image, $x_i$, has the same expected classification error at each $DP_k$. This is a strong assumption, and can lead to suboptimal performance, e.g., when DPs specialize to different images (see Section~\ref{sec:specialization}).
There are a number of ways to address this issue.
One way would be to enhance Algorithm~\ref{alg_alloc} with an additional heuristic. In particular, if the classification probability of an image which is assigned to a DP is higher than a threshold, it will never be assigned to another DP, irrespective of the available budget.

An alternative approach would be to train a gating network to decide which images should be assigned to each DP.
This is similar to how images are assigned to different models in a Mixture of Experts (MoE). The difference in our setting is that the number of images that should be assigned to each model is determined by solving program~\ref{eq20}; MoE usually don't have any constraint on how many images can be classified by each model.
An additional benefit of using a gating network is that members of an ensemble do not have to be executed sequentially, which is another limitation of our current content-sensitive strategy.

Finally, using multiple models simultaneously, as in the case of the multi-resolution ensemble, may be prohibitive in some settings due to memory constraints.
In such settings, natively adjustable models, like TNet, may be more suitable.
\section{Conclusion}
\label{sec:conclusion}
We formulated budgeted image classification as a resource allocation integer program, which is NP-Hard. We offered two approximate solutions, a content-agnostic, and a content-sensitive one.
We tested both strategies on satellite images (fMoW), by using a multi-resolution ensemble, and a hard-attention model.
We experimentally showed that content-sensitive heuristics can offer significant accuracy benefits,
and we theoretically analyzed conditions that must be satisfied by decision points to be suitable for budgeted classification.
Finally, we identified limitations to our strategies, including their inability to handle decision points that specialize to different modes of the data distribution, and we discussed potential solutions as future research directions.
{
    \small
    \bibliographystyle{ieeenat_fullname}
    \bibliography{main}

@article{goyal2017accurate,
  title={Accurate, large minibatch sgd: Training imagenet in 1 hour},
  author={Goyal, Priya and Doll{\'a}r, Piotr and Girshick, Ross and Noordhuis, Pieter and Wesolowski, Lukasz and Kyrola, Aapo and Tulloch, Andrew and Jia, Yangqing and He, Kaiming},
  journal={arXiv preprint arXiv:1706.02677},
  year={2017}
}

@article{howard2017mobilenets,
  title={Mobilenets: Efficient convolutional neural networks for mobile vision applications},
  author={Howard, Andrew G and Zhu, Menglong and Chen, Bo and Kalenichenko, Dmitry and Wang, Weijun and Weyand, Tobias and Andreetto, Marco and Adam, Hartwig},
  journal={arXiv preprint arXiv:1704.04861},
  year={2017}
}

@inproceedings{howard2019searching,
  title={Searching for mobilenetv3},
  author={Howard, Andrew and Sandler, Mark and Chu, Grace and Chen, Liang-Chieh and Chen, Bo and Tan, Mingxing and Wang, Weijun and Zhu, Yukun and Pang, Ruoming and Vasudevan, Vijay and others},
  booktitle={Proceedings of the IEEE/CVF International Conference on Computer Vision},
  pages={1314--1324},
  year={2019}
}

@inproceedings{tan2019efficientnet,
  title={Efficientnet: Rethinking model scaling for convolutional neural networks},
  author={Tan, Mingxing and Le, Quoc},
  booktitle={International Conference on Machine Learning},
  pages={6105--6114},
  year={2019},
  organization={PMLR}
}

@article{tan2021efficientnetv2,
  title={Efficientnetv2: Smaller models and faster training},
  author={Tan, Mingxing and Le, Quoc V},
  journal={arXiv preprint arXiv:2104.00298},
  year={2021}
}

@article{bello2021revisiting,
  title={Revisiting resnets: Improved training and scaling strategies},
  author={Bello, Irwan and Fedus, William and Du, Xianzhi and Cubuk, Ekin D and Srinivas, Aravind and Lin, Tsung-Yi and Shlens, Jonathon and Zoph, Barret},
  journal={arXiv preprint arXiv:2103.07579},
  year={2021}
}

@inproceedings{touvron2021training,
  title={Training data-efficient image transformers \& distillation through attention},
  author={Touvron, Hugo and Cord, Matthieu and Douze, Matthijs and Massa, Francisco and Sablayrolles, Alexandre and J{\'e}gou, Herv{\'e}},
  booktitle={International Conference on Machine Learning},
  pages={10347--10357},
  year={2021},
  organization={PMLR}
}

@article{strubell2019energy,
  title={Energy and policy considerations for deep learning in NLP},
  author={Strubell, Emma and Ganesh, Ananya and McCallum, Andrew},
  journal={arXiv preprint arXiv:1906.02243},
  year={2019}
}

@inproceedings{bolukbasi2017adaptive,
  title={Adaptive neural networks for efficient inference},
  author={Bolukbasi, Tolga and Wang, Joseph and Dekel, Ofer and Saligrama, Venkatesh},
  booktitle={International Conference on Machine Learning},
  pages={527--536},
  year={2017},
  organization={PMLR}
}

@inproceedings{ruiz2019adaptative,
  title={Adaptative inference cost with convolutional neural mixture models},
  author={Ruiz, Adria and Verbeek, Jakob},
  booktitle={Proceedings of the IEEE/CVF International Conference on Computer Vision},
  pages={1872--1881},
  year={2019}
}

@article{huang2017multi,
  title={Multi-scale dense networks for resource efficient image classification},
  author={Huang, Gao and Chen, Danlu and Li, Tianhong and Wu, Felix and van der Maaten, Laurens and Weinberger, Kilian Q},
  journal={arXiv preprint arXiv:1703.09844},
  year={2017}
}

@inproceedings{teerapittayanon2016branchynet,
  title={Branchynet: Fast inference via early exiting from deep neural networks},
  author={Teerapittayanon, Surat and McDanel, Bradley and Kung, Hsiang-Tsung},
  booktitle={2016 23rd International Conference on Pattern Recognition (ICPR)},
  pages={2464--2469},
  year={2016},
  organization={IEEE}
}

@inproceedings{li2019improved,
  title={Improved techniques for training adaptive deep networks},
  author={Li, Hao and Zhang, Hong and Qi, Xiaojuan and Yang, Ruigang and Huang, Gao},
  booktitle={Proceedings of the IEEE/CVF International Conference on Computer Vision},
  pages={1891--1900},
  year={2019}
}

@inproceedings{figurnov2017spatially,
  title={Spatially adaptive computation time for residual networks},
  author={Figurnov, Michael and Collins, Maxwell D and Zhu, Yukun and Zhang, Li and Huang, Jonathan and Vetrov, Dmitry and Salakhutdinov, Ruslan},
  booktitle={Proceedings of the IEEE Conference on Computer Vision and Pattern Recognition},
  pages={1039--1048},
  year={2017}
}

@inproceedings{veit2018convolutional,
  title={Convolutional networks with adaptive inference graphs},
  author={Veit, Andreas and Belongie, Serge},
  booktitle={Proceedings of the European Conference on Computer Vision (ECCV)},
  pages={3--18},
  year={2018}
}

@inproceedings{wang2018skipnet,
  title={Skipnet: Learning dynamic routing in convolutional networks},
  author={Wang, Xin and Yu, Fisher and Dou, Zi-Yi and Darrell, Trevor and Gonzalez, Joseph E},
  booktitle={Proceedings of the European Conference on Computer Vision (ECCV)},
  pages={409--424},
  year={2018}
}

@article{yu2018slimmable,
  title={Slimmable neural networks},
  author={Yu, Jiahui and Yang, Linjie and Xu, Ning and Yang, Jianchao and Huang, Thomas},
  journal={arXiv preprint arXiv:1812.08928},
  year={2018}
}

@article{gao2018dynamic,
  title={Dynamic channel pruning: Feature boosting and suppression},
  author={Gao, Xitong and Zhao, Yiren and Dudziak, {\L}ukasz and Mullins, Robert and Xu, Cheng-zhong},
  journal={arXiv preprint arXiv:1810.05331},
  year={2018}
}

@inproceedings{yang2020resolution,
  title={Resolution adaptive networks for efficient inference},
  author={Yang, Le and Han, Yizeng and Chen, Xi and Song, Shiji and Dai, Jifeng and Huang, Gao},
  booktitle={Proceedings of the IEEE/CVF Conference on Computer Vision and Pattern Recognition},
  pages={2369--2378},
  year={2020}
}

@article{wang2021not,
  title={Not All Images are Worth 16x16 Words: Dynamic Vision Transformers with Adaptive Sequence Length},
  author={Wang, Yulin and Huang, Rui and Song, Shiji and Huang, Zeyi and Huang, Gao},
  journal={arXiv preprint arXiv:2105.15075},
  year={2021}
}

@article{odena2017changing,
  title={Changing model behavior at test-time using reinforcement learning},
  author={Odena, Augustus and Lawson, Dieterich and Olah, Christopher},
  journal={arXiv preprint arXiv:1702.07780},
  year={2017}
}

@article{shazeer2017outrageously,
  title={Outrageously large neural networks: The sparsely-gated mixture-of-experts layer},
  author={Shazeer, Noam and Mirhoseini, Azalia and Maziarz, Krzysztof and Davis, Andy and Le, Quoc and Hinton, Geoffrey and Dean, Jeff},
  journal={arXiv preprint arXiv:1701.06538},
  year={2017}
}

@article{graves2016adaptive,
  title={Adaptive computation time for recurrent neural networks},
  author={Graves, Alex},
  journal={arXiv preprint arXiv:1603.08983},
  year={2016}
}

@inproceedings{verelst2020dynamic,
  title={Dynamic convolutions: Exploiting spatial sparsity for faster inference},
  author={Verelst, Thomas and Tuytelaars, Tinne},
  booktitle={Proceedings of the IEEE/CVF Conference on Computer Vision and Pattern Recognition},
  pages={2320--2329},
  year={2020}
}

@inproceedings{li2017dynamic,
  title={Dynamic computational time for visual attention},
  author={Li, Zhichao and Yang, Yi and Liu, Xiao and Zhou, Feng and Wen, Shilei and Xu, Wei},
  booktitle={Proceedings of the IEEE International Conference on Computer Vision Workshops},
  pages={1199--1209},
  year={2017}
}

@article{wang2020glance,
  title={Glance and focus: a dynamic approach to reducing spatial redundancy in image classification},
  author={Wang, Yulin and Lv, Kangchen and Huang, Rui and Song, Shiji and Yang, Le and Huang, Gao},
  journal={arXiv preprint arXiv:2010.05300},
  year={2020}
}

@article{papadopoulos2021hard,
  title={Hard-attention for scalable image classification},
  author={Papadopoulos, Athanasios and Korus, Pawel and Memon, Nasir},
  journal={Advances in Neural Information Processing Systems},
  volume={34},
  pages={14694--14707},
  year={2021}
}

@article{brown2020language,
  title={Language models are few-shot learners},
  author={Brown, Tom and Mann, Benjamin and Ryder, Nick and Subbiah, Melanie and Kaplan, Jared D and Dhariwal, Prafulla and Neelakantan, Arvind and Shyam, Pranav and Sastry, Girish and Askell, Amanda and others},
  journal={Advances in neural information processing systems},
  volume={33},
  pages={1877--1901},
  year={2020}
}

@article{touvron2023llama,
  title={Llama: Open and efficient foundation language models},
  author={Touvron, Hugo and Lavril, Thibaut and Izacard, Gautier and Martinet, Xavier and Lachaux, Marie-Anne and Lacroix, Timoth{\'e}e and Rozi{\`e}re, Baptiste and Goyal, Naman and Hambro, Eric and Azhar, Faisal and others},
  journal={arXiv preprint arXiv:2302.13971},
  year={2023}
}

@inproceedings{fmow2018,
  title={Functional Map of the World},
  author={Christie, Gordon and Fendley, Neil and Wilson, James and Mukherjee, Ryan},
  booktitle={CVPR},
  year={2018}
}
}


\end{document}